\documentclass{article}

\usepackage[nonatbib, preprint]{neurips_2022}

\usepackage[
    natbib=true,
    style=numeric,
    sorting=none
]{biblatex}
\usepackage[utf8]{inputenc} % allow utf-8 input
\usepackage[T1]{fontenc}    % use 8-bit T1 fonts
\usepackage{hyperref}       % hyperlinks
\usepackage{url}            % simple URL typesetting
\usepackage{booktabs}       % professional-quality tables
\usepackage{amsfonts}       % blackboard math symbols
\usepackage{nicefrac}       % compact symbols for 1/2, etc.
\usepackage{microtype}      % microtypography
\usepackage{xcolor}         % colors
\usepackage{float}          % suppressing table floating
\usepackage{graphicx}       % images
\usepackage{amsmath}        % math
\usepackage{amssymb}        % math symbols
\usepackage{siunitx}        % scientific units
\usepackage{multirow}       % multirow support
\usepackage{subcaption}     % subcaptions

\DeclareSIUnit\angstrom{\text {Å}}

\title{\textsc{EGR}: Equivariant Graph Refinement and Assessment of 3D Protein Complex Structures}

\author{%
  Alex Morehead, Xiao Chen, Tianqi Wu, Jian Liu, Jianlin Cheng \\
  Department of Electrical Engineering \& Computer Science \\
  University of Missouri\\
  Columbia, MO 65211 \\
  \texttt{\{acmwhb, xcbh6, wuti, jl4mc, chengji\}@umsystem.edu} \\
}

\begin{document}

\maketitle

\begin{abstract}
  Protein complexes are macromolecules essential to the functioning and well-being of all living organisms. As the structure of a protein complex, in particular its region of interaction between multiple protein subunits (i.e., chains), has a notable influence on the biological function of the complex, computational methods that can quickly and effectively be used to refine and assess the quality of a protein complex's 3D structure can directly be used within a drug discovery pipeline to accelerate the development of new therapeutics and improve the efficacy of future vaccines. In this work, we introduce the Equivariant Graph Refiner (\textsc{EGR}), a novel E(3)-equivariant graph neural network (GNN) for multi-task structure refinement and assessment of protein complexes. Our experiments on new, diverse protein complex datasets, all of which we make publicly available in this work, demonstrate the state-of-the-art effectiveness of \textsc{EGR} for atomistic refinement and assessment of protein complexes and outline directions for future work in the field. In doing so, we establish a baseline for future studies in macromolecular refinement and structure analysis.\footnote{Inference code as well as pre-trained models are available at \href{https://github.com/BioinfoMachineLearning/DeepRefine}{\texttt{https://github.com/BioinfoMachineLearning/DeepRefine}}}
\end{abstract}

\section{Introduction}
Protein complexes, and their structures in particular, underpin many important biological functions in living organisms \cite{doi:10.1073/pnas.2032324100}. With an enhanced understanding of how protein chains interact to form complexes, fundamental research in fields such as drug discovery and materials science is likely to accelerate considerably \cite{bahadur2008interface}. For example, to obtain the structure of a complex thought to interact with a candidate compound, in a typical drug discovery pipeline one must first either analytically derive its structure using powerful yet time and resource-intensive techniques such as X-ray crystallography and nuclear magnetic resonance spectroscopy \cite{WALZTHOENI2013252} or instead directly predict the structure using computational methods \cite{leelananda2016computational}.

The benefits offered by using purely computational approaches for structure determination are numerous, such as significantly accelerating the quantity and speed at which one can obtain 3D structures for downstream studies. Such tools, which for single protein chains have advanced considerably in the last several years \cite{lensink2021prediction}, have been used to accelerate the analysis of protein function at genomic scales \cite{gao2021high} and have inspired the development of new methods designed to predict interactions between protein chains \cite{morehead2021dips, morehead2021geometric} and to assess the quality of 3D protein structures \cite{chen20223d}. However, these methods are currently less reliable for modeling protein complexes \cite{tunyasuvunakool2022prospects}.

In such cases, researchers then turn to structure refinement methods that aim to improve the quality of initial protein structures via either statistical techniques or learning-based methods \cite{verburgt2022benchmarking}. Computational tools designed to refine and assess the structure of protein complexes, especially low-quality structures, can simultaneously increase the available quality of such structures while offering researchers enhanced insight into the variety of molecular conformations and functions into which a complex can resolve. Nonetheless, refinement of 3D protein structures is a difficult task, with many challenges arising from the size of the search space in which a higher-quality structure can be found and the physical constraints that must be respected by an atomic system to form a realistic protein structure. Consequently, existing methods for complex refinement are driven by molecular dynamics and relaxation protocols, which rely on expert knowledge embedded in the system through the design of such protocols. The limits of such methods have previously been explored \cite{verburgt2022benchmarking} and, as such, there is an increasing level of interest to see if learning-based methods (e.g., deep learning (DL) models) can be used to refine complex structures.

\begin{figure}
    \centering
    \includegraphics[width=\textwidth]{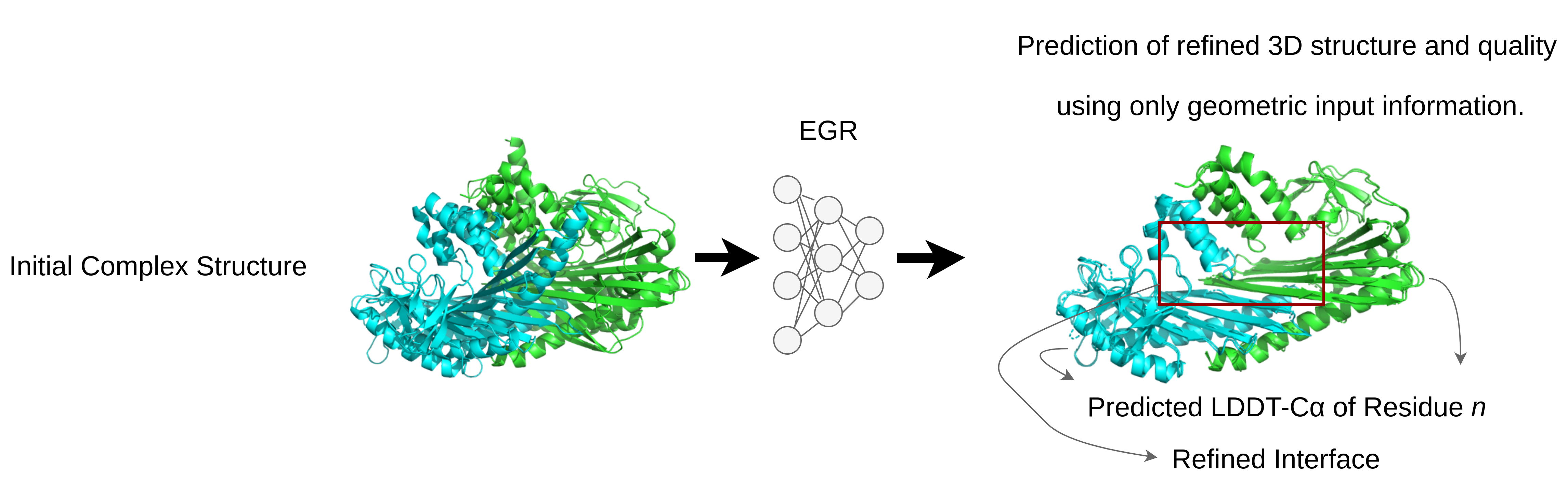}
    \caption{An overview of the structure refinement and assessment problem addressed by \textsc{EGR}.}
    \label{fig:problem_overview}
\end{figure}

Here, we introduce \textsc{EGR}, a novel E(3)-equivariant graph deep learning model for multi-task structure refinement and quality assessment of protein complexes - Figure \ref{fig:problem_overview}. Notably, we exploit equivariant graph neural networks (EGNNs) \cite{satorras2021n} and careful input regularization to make a direct prediction of the refined structure for an initial protein complex as well as the per-residue quality of the refined structure. Given that our method requires only a single forward pass to finalize its predictions, we achieve \textbf{significant speed-ups in inference time} compared to traditional refinement software solutions. Moreover, our model refines the positions of \textit{all} atoms in an input protein, making it the first of its kind in DL-based complex structure refinement.

\section{Related Work}
We now proceed to describe prior works relevant to DL-based structure refinement and assessment.\par

\textbf{Biomolecular structure prediction.} Until very recently, the conventional means of determining a molecule's 3D structure would involve costly and time-consuming physical trials performed by experimental scientists \cite{bill2011overcoming}. However, using fast computational inference, new DL methods have now made it possible to determine the structures of proteins and other biomolecules in a matter of minutes rather than weeks or months \cite{jumper2021highly, baek2021accurate}. Such methods have promoted the widespread adoption of structure prediction software \cite{varadi2022alphafold, cramer2021alphafold2} and have inspired several new works in DL-driven structure prediction \cite{doi:10.1126/science.abe5650, stark2022equibind}.\par

\textbf{Protein representation learning (for DL-based molecular modeling).} Proteins can be represented in many different forms. For example, representing proteins by their amino acid sequences has been shown to provide powerful structural information by comparing sequences to each other \cite{jumper2021highly} and extracting rich unsupervised learning representations using self-attention Transformers \cite{rao2021transformer}. From a geometric viewpoint, several works have aimed to encode protein structural priors directly within neural network architectures to model proteins hierarchically \cite{somnath2021multi, hermosilla2020intrinsic}, as computationally-efficient point clouds \cite{gainza2020deciphering, sverrisson2021fast}, or as k-nearest neighbors (k-NN) geometric graphs \cite{jing2020learning, jin2021iterative} for tasks such as protein function prediction \cite{gligorijevic2021structure}, protein model quality assessment \cite{eismann2020protein}, and protein interaction region prediction \cite{dai2021protein}.\par

\textbf{Applications of structure prediction methods.} Markedly, structure prediction models have been applied to predict the structures of all proteins in the human proteome and have aided in efforts to better understand the mechanisms underlying disordered proteins \cite{ruff2021alphafold}. Moreover, DL-based structure prediction methods have accelerated the discovery of promising drug candidates for therapeutics research \cite{ren2022alphafold} and have enabled fast virtual drug screening \cite{gniewek2021learning} and materials discovery \cite{oganov2019structure} at scale.\par

\textbf{Deep learning for protein-protein docking.} Recent advancements in geometric deep learning have supported the development of new DL-based models for rigid body protein-protein docking \cite{sverrisson2022physics}, as well as docking in a single-shot setting \cite{ganea2021independent}. In addition, previous DL models for single-chain protein structure prediction have been repurposed to predict protein complex structures \cite{evans2021protein} and refine the geometry of side-chain atoms \cite{jindal2021side}.\par

\textbf{Deep learning for protein structure refinement.} DL models have also been used to guide the refinement of residue positions within tertiary protein structures \cite{hiranuma2021improved} or predict refined residue positions using indirect target values such as inter-residue distances \cite{10.1093/nar/gkab361, jing2021fast}. Unfortunately, such methods, following refinement, require all-atom restoration procedures as a post-processing step to recover the positions of backbone and side-chain atoms. Promisingly, \cite{Wu2022.05.06.490934} have begun exploring the use of DL models for all-atom refinement of protein tertiary structures, however, this method has demonstrated success solely for tertiary structures due to its high computational memory complexity.\par

\textbf{Deep learning for protein structure quality assessment.} As DL-driven structure prediction methods have matured, new methods for quality assessment (QA) of protein structures have also been developed to facilitate automatic ranking of protein structures. In particular, 2D convolutional neural networks (CNNs) \cite{chen2022distema}, 3D CNNs \cite{pages2019protein}, and GNNs \cite{sanyal2020proteingcn, baldassarre2021graphqa, chen20223d} have recently been adopted for structure ranking.\par

\textbf{Incorporating symmetries in GNNs.} Embuing neural networks with inductive priors has a longstanding history in the field \cite{battaglia2018relational, feinman2018learning, neyshabur2014search}. Priors of particular importance for 3D structure modeling include rotation \cite{marcos2017rotation} and translation \cite{thomas2018tensor} equivariance. Such priors form the basis of 3D Euclidean transformations that can now directly be found within neural network layers \cite{cohen2016group, fuchs2020se, satorras2021n, brandstetter2022geometric, batzner20223} to increase networks' data efficiency and generalization capabilities \cite{he2021efficient, bulusu2021generalization}. Our work follows that of \cite{satorras2021n} to incorporate E(3)-equivariance in our message passing neural network for 3D structure refinement and quality assessment. However, we go beyond this method by adding skip connections shown to improve gradient flow after position updates, regularizing node input features according to geometric priors for 3D molecules, and employing loss functions for refinement and quality assessment in a semi-supervised manner, respectively.\par

\textbf{Contributions.}\\ Our work builds upon prior works by making the following contributions:

\begin{itemize}
\item We provide the \textit{first} example of applying deep learning to the task of all-atom refinement of protein complex structures.
\item We provide the \textit{first} example of applying equivariant graph message passing to the multi-task setting of refining and assessing protein complex structures concurrently.
\item We introduce the new semi-supervised \textsc{EGR} model, showcasing its effective use in improving the structure of interface regions between protein chains and estimating its confidence in such improvements.
\end{itemize}

\section{\textsc{EGR} Model}
\label{sec:egr_model}

\begin{figure}
    \centering
    \includegraphics[width=\textwidth]{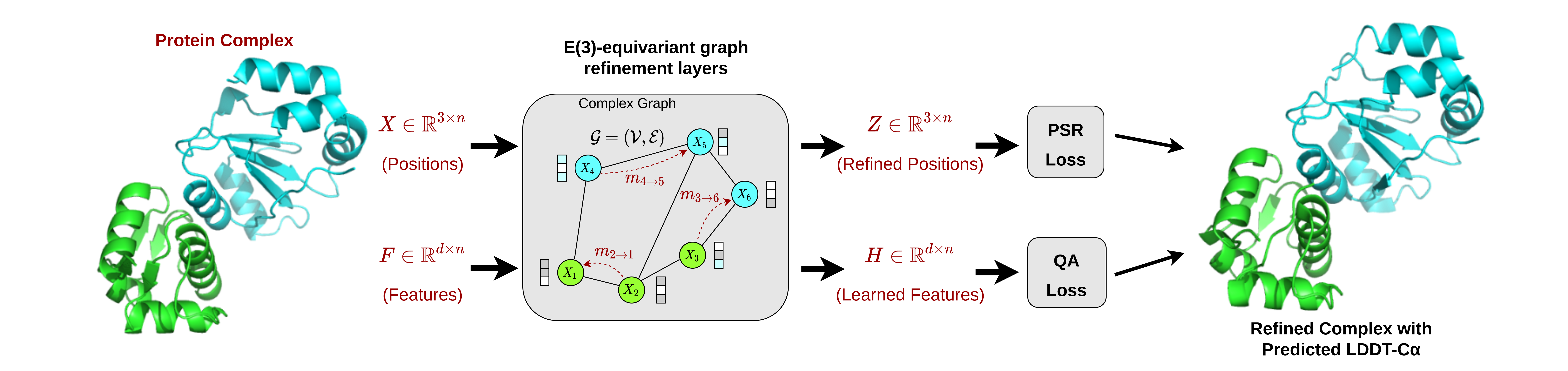}
    \caption{\textsc{EGR} model architecture.}
    \label{fig:egr_architecture}
\end{figure}

We now turn to describe the \textsc{EGR} model, which is shown in Figure \ref{fig:problem_overview} and detailed in Figure \ref{fig:egr_architecture}. The \textsc{EGR} model receives as its input a single 3D protein complex graph representing an initial decoy structure for a given protein target. As such, all chains in the complex are modeled within the same homogeneous input graph structure. As described in Appendix \ref{sec:appendix_b}, we differentiate pairs of atoms belonging to different chains using a one-hot encoded edge feature to initialize the \textsc{EGR} model with knowledge of the complex's separate chains. In doing so, we allow for structural flexibility of each protein chain in our work, without assuming any rigid conformations within a complex. Notably, our model does this while being trained using only geometric information obtained from each input protein, as it does not make use of \textit{any} coevolutionary or hand-crafted features for the task at hand.\par

\textbf{K-NN graph representation.} The input protein complex graphs for our model are constructed using a spatial k-NN algorithm, where a graph $\mathcal{G = (V, E)}$ treats the protein's atoms as its nodes and denotes each atom's initial features as $\mathbf{F} \in \mathbb{R}^{d \times n}$ (e.g., atom type) and its positions in $\mathbb{R}^{3}$ as $\mathbf{X} \in \mathbb{R}^{3 \times n}$. The nodes in $\mathcal{G}$ are connected to their 20 closest node neighbors in $\mathbb{R}^{3}$, yielding a total of $n \times 20$ edges within $\mathcal{G}$. In Appendix \ref{sec:appendix_b}, we describe the node features $\mathbf{F}$ in further detail.\par

\textbf{E(3)-equivariant transformations.} In this work, we argue that E(3)-equivariance offers a suitable inductive prior for modeling 3D protein structures. We make this point based primarily on the computational efficiency of E(3)-equivariant networks today. As such, for data efficiency and generalization capabilities, we then turn to designing a neural network capable of capturing within its hidden layers E(3)-transformations of any input protein, especially since for this task we are left to train on a relatively small number of input examples. Towards this end, we propose the \textit{Equivariant Graph Refiner} (\textsc{EGR}) model, which combines insights from Equivariant Graph Neural Networks \cite{satorras2021n}, EquiDock \cite{ganea2021independent}, and EquiBind \cite{stark2022equibind}. The \textsc{EGR} model learns to transform node features and node positions in $\mathbb{R}^{3}$ to perform graph message passing \textit{across} each input complex graph. Implicitly, this means our model is exchanging information between protein chains in a complex. More formally, \textit{EGR}($\mathbf{X}$, $\mathbf{F}$) = $\mathbf{Z} \in \mathbb{R}^{3 \times n}, \mathbf{H} \in \mathbb{R}^{d \times n}$, where $\mathbf{H}$ represents node embeddings and $\mathbf{Z}$ represents transformed node coordinates. We note that, in practical terms, \textsc{EGR} layers can be stacked sequentially in an E(3)-equivariant manner, such that 3D translations or rotations of an input graph will be reflected in the output of any \textsc{EGR} layer. Namely, for an arbitrary translation vector $\mathbf{b} \in \mathbb{R}^{3}$ and orthogonal matrix $\mathbf{U} \in \mathbf{SO}(3)$, \textit{EGR}($\mathbf{UX} + \mathbf{b}, \mathbf{F}$) = $\textbf{UZ} + \textbf{b}, \mathbf{H}$. Subsequently, our definition of a single \textsc{EGR} layer is:\par

\begin{equation}
    \mathbf{m}_{j \rightarrow i} = \varphi^{e}(\mathbf{h}_{i}^{(l)}, \mathbf{h}_{j}^{(l)}, \mathbf{f}_{j \rightarrow i}, \| \mathbf{x}_{i}^{(l)} - \mathbf{x}_{j}^{(l)} \|^{2}), \forall(i,j) \in \mathcal{E}
\end{equation}

\begin{equation}
    \label{eq:a_2}
    \mathbf{x}_{i}^{(l + 1)} = \alpha\mathbf{x}_{i}^{(0)} + (1 - \alpha)\mathbf{x}_{i}^{(l)} + \frac{1}{| \mathcal{N}(i) |} \sum_{j \in \mathcal{N}(i)} \frac{(\mathbf{x}_{i}^{(l)} - \mathbf{x}_{j}^{(l)})}{\| \mathbf{x}_{i}^{(l)} - \mathbf{x}_{j}^{(l)} \| + \mathcal{C}}\ \varphi^{x}(\mathbf{m}_{j \rightarrow i})
\end{equation}

\begin{equation}
    \mathbf{m}_{i} = \frac{1}{| \mathcal{N}(i) |} \sum_{j \in \mathcal{N}(i)} \mathbf{m}_{j \rightarrow i}, \forall i \in \mathcal{V}
\end{equation}

\begin{equation}
    \mathbf{h}_{i}^{(l + 1)} = \beta \cdot \varphi^{h}(\mathbf{h}_{i}^{(l)}, \mathbf{m}_{i}, \mathbf{a}_{i}, \mathbf{f}_{i}) + (1 - \beta) \cdot \mathbf{h}_{i}^{(l)}, \forall i \in \mathcal{V},
\end{equation}

where $\alpha$ and $\beta$, respectively, represent scalar coordinates-wise and node representations-wise skip connection strengths; $\mathcal{N}(i)$ denotes the graph neighbors of node $i$; $\mathbf{a}_{i}$ are optional SE(3)-invariant attention coefficients \cite{ganea2021independent} derived from $\mathbf{H}$; $\mathcal{C}$ is a normalization constant that we set to $\mathcal{C} = 1$; and the remaining $\varphi$ functions are represented as shallow neural networks, with $\varphi^{x}$ producing a single scalar and all others a $d$-dimensional vector. Our rationale for normalizing relative positional displacements (i.e., $\mathbf{x}_{i}^{(l)} - \mathbf{x}_{j}^{(l)}$) by their norm, in a similar manner as \cite{satorras2021f}, is to keep the updated coordinates of each \textsc{EGR} layer from exploding in value when aggregating updates from each successive layer. The term $\mathcal{C}$ is also included here to ensure that Equation \ref{eq:a_2} is still differentiable and to stabilize the \textsc{EGR} model's training when using $\mathcal{C} = 1$. We note that in our studies here, we compute $a_{i}$ from $\mathbf{H}$ using a variant of the Linear Attention Transformer architecture \cite{vaswani2017attention, DBLP:journals/corr/abs-1812-01243, wang2021}, as described in Appendix \ref{sec:appendix_b}.

\subsection{Refining initial coordinates} As mentioned previously, each \textsc{EGR} layer predicts transformed coordinates $\mathbf{Z}$. The output of our model's last \textsc{EGR} layer, $\mathbf{Z}^{L}$, then serves as our estimate of the input complex's refined coordinates, $\mathbf{X}'$ (i.e., $\mathbf{X}' = \mathbf{Z}^{L}$).\par

\textbf{Refinement loss.} To guide the \textsc{EGR} model towards a refined molecular conformation, we use the Huber loss \cite{meyer2021alternative} between the \textsc{EGR} model's refined coordinates $\mathbf{X}'$ and the ground truth coordinates $\mathbf{X}^{*}$. Formally, the \textsc{EGR} model's loss for refinement (i.e., PSR loss) is:

\[
\mathcal{L}_{PSR}^{i} = \begin{cases}
            \frac{1}{2} (x_{i}' - x_{i}^{*})^2, & \text{if } |x_{i}' - x_{i}^{*}| < \delta, \\
            \delta * (|x_{i}' - x_{i}^{*}| - \frac{1}{2} * \delta) & \text{otherwise}
          \end{cases}
\]

\begin{equation}
    \mathcal{L}_{PSR} = \frac{1}{| \mathbf{P} |} \sum_{i \in \mathbf{P}} \mathcal{L}_{PSR}^{i},
\end{equation}

where $\mathbf{x}_{i}'$ are the model's predicted coordinates for node $i$, $\mathbf{x}_{i}^{*}$ are the ground truth coordinates for node $i$, and $\mathbf{P}$ is the set of indices of atoms for which ground truth coordinates exist in the corresponding native structure. Additionally, we let $\delta = 1.0$ in the context of this study. In the context of refinement, outlier coordinate predictions may occur at any stage during model training, so we must take care to ensure our network's gradients do not explode or saturate inadvertently. As such, we chose this refinement loss function to make it less sensitive to outliers than a mean squared error (MSE) criterion and for it to be smoother near zero compared to a mean absolute error criterion.\par

\textbf{Positional reconstruction.} Immediately before training, we inject random Gaussian noise into the initial state of our node positions. This corresponds to a small corruption of the location of each atom in an input protein. This methodology follows closely after the Noisy Nodes methodology from \cite{godwin2021very} and, as such, may facilitate better learning over the manifold of ground truth structures. We study the behavior of the \textsc{EGR} model with and without positional corruption in Section \ref{sec:appendix_a}.\par

\subsection{Assessing structure quality} Following refinement of initial node coordinates, the \textsc{EGR} model also predicts new embeddings for each node, $\mathbf{H}$. Intuitively, we consider $\mathbf{H}$ to be the model's knowledge of its predicted molecular structure as defined by $\mathbf{X}'$. As such, using $\mathbf{H}$, we train an additional head of the network to predict an estimate of the structural accuracy of $\mathbf{X}'$.

\textbf{Quality assessment loss.} We formulate assessment of a predicted structure's quality as a node regression task. Under this framework, we employ an MSE loss to train the second head of the network to predict the carbon-alpha (C$\alpha$) atom local Distance Difference Test (LDDT-C$\alpha$) value \cite{mariani2013lddt} of each residue in the complex graph. Formally, the \textsc{EGR} model's loss for quality assessment (i.e., QA loss) is:

\begin{equation}
    \mathcal{L}_{QA} = \frac{1}{| \mathbf{C} |} \sum_{i \in \mathbf{C}} \| \mathbf{q}_{i}' - \mathbf{q}_{i}^{*} \|^{2},
\end{equation}

where $\mathbf{q}_{i}'$ is the model's predicted LDDT-C$\alpha$ for node $i$, $\mathbf{q}_{i}^{*}$ is the ground truth LDDT-C$\alpha$ for node $i$, and $\mathbf{C}$ is the set of C$\alpha$ atom indices in the input graph. As demonstrated by \cite{jumper2021highly}, models trained to predict such a quantity can produce reasonable accuracy estimates for predicted structures, taking one step towards making the \textsc{EGR} model more interpretable for its users.

\subsection{Learning paradigm} As mentioned previously, not all atoms in an input decoy structure have a corresponding position in the ground truth structure. Moreover, since during training we can only calculate the ground truth LDDT labels for the input graph's C$\alpha$ atoms, the \textsc{EGR} model jointly learns to predict refined structures and LDDT-C$\alpha$ scores in a \textit{semi-supervised} end-to-end manner. \par

\section{Experiments}
\label{sec:experiments}

\subsection{Data}
\label{sec:data}
For the task of complex structure refinement and assessment, we introduce new Protein Structure Refinement (PSR) datasets comprised of 46,174 homomeric and heteromeric protein complex structural decoys, corresponding to 3,498 unique protein targets. The proteins comprising these datasets (i.e., PSR-Dockground, PSR-DeepHomo, and PSR-EVCoupling) are originally derived from the Dockground \cite{kundrotas2018dockground}, DeepHomo \cite{yan2021accurate}, and EVCoupling \cite{hopf2019evcouplings} datasets, respectively. The Dockground dataset provides decoy structures of various structural error ranges for each protein target, so we directly include its 1-6 \si{\angstrom} decoys and their ground truth structures in our PSR dataset. As the DeepHomo and EVCoupling datasets only provide ground truth structures for each of their protein targets, to generate each target's decoy structures for refinement, we subsequently use AlphaFold-Multimer \cite{evans2021protein} to predict and select the best decoy structure for each of these datasets' protein targets. Using TM-score \cite{zhang2004scoring}, we then uniquely superimpose all ground truth target structures corresponding to each decoy structure to ensure the coordinate systems of both decoy and target structures are aligned before model training. To support the reproducibility of our work, we make our PSR datasets publicly available along with standardized training, validation, and testing splits of the input proteins.\par

\textbf{Motivation for introducing new datasets.} To the best of our knowledge, our work is the first to introduce cross-validation datasets curated for structural refinement of protein complexes, datasets that we have intentionally designed to be amenable to machine learning methods. In constructing these datasets, we sought to make use of existing protein data sources to maximize their size and structural scope. Notably, since the average structural quality of AlphaFold-Multimer for DeepHomo and EVCoupling proteins is reasonably high, we included proteins from the Dockground dataset to diversify the overall quality of structures in our PSR dataset.\par

\textbf{Overlap Reduction.} Our constituent PSR datasets are initially combined using random population-proportionate sampling according to a protein's type and dataset of origin (i.e., Dockground-Heteromer, DeepHomo-Homomer, and EVCoupling-Heteromer). Using MMseqs2 \cite{steinegger2017mmseqs2}, these proportionate splits are then filtered such that no split contains a protein chain with more than 30\% sequence similarity to any chain in another split, across all datasets employed in this study. This filtering technique, among others, was chosen to satisfy multiple practical criteria for tasks in this domain: assurance that models (1) cross-validated on such splits do not overfit w.r.t. homology \cite{10.1093/bioinformatics/btm006} or structural similarity \cite{koehl2002sequence} between proteins and (2) will be trained on a sufficient number of input proteins to enable generalization to unseen data.\par

\textbf{Test Datasets.} The first test dataset we use to evaluate models performing best on our PSR dataset's validation split is comprised of complexes originally held out from our PSR dataset. As such, this PSR test dataset is comprised of Dockground, DeepHomo, and EVCoupling proteins that have previously been filtered according to 30\% sequence similarity. Following the work of \cite{evans2021protein}, we then adopt the heteromeric Benchmark 2 dataset for blind evaluation of our model's generalization for the task of refinement. Additionally, to score our model's ability to estimate the quality of a protein complex, we curated a new M4S dataset for complex quality assessment. This dataset consists of randomly-sampled heteromeric protein complexes available in the Protein Data Bank (PDB), where each of the 11 selected complexes was first subjected to 30\% sequence identity filtering w.r.t. all of our other cross-validation dataset splits, yielding 1,160 decoys across 11 targets for QA benchmarking.\par

\subsection{Evaluation Setup}
\label{sec:evaluation_setup}

\textbf{Baselines.} Modeller \cite{webb2016comparative} is a classical refinement program for 3D protein structures, with support for modeling protein complexes in particular. We also include a version of \textsc{EGR} trained for only C$\alpha$ atom refinement and assessment (i.e., \textsc{EGR}-C$\alpha$) to evaluate Modeller's ability to reconstruct all-atom structures from the output C$\alpha$ atom coordinates of \textsc{EGR}-C$\alpha$. GalaxyRefineComplex \cite{heo2016galaxyrefinecomplex} is another popular refinement protocol, one specifically designed for refining structural interfaces between chains in a protein complex. Similarly, we also include GNNRefine \cite{jing2021fast}, which uses GNN-based distance predictions to drive tertiary structural refinement with PyRosetta \cite{chaudhury2010pyrosetta}. We note that we are only able to include GalaxyRefineComplex and GNNRefine's results on our smaller Benchmark 2 test dataset, as their extraordinarily high refinement runtimes (e.g., 1,200 seconds per decoy with 16 CPU threads) prevent us from evaluating their performance in a reasonable amount of time on our full PSR test dataset consisting of over 5,000 complexes. For our remaining all-atom baselines, we selected two recent geometric deep learning methods with which to compare the \textsc{EGR} model's performance for all-atom refinement and QA. These methods include SE(3)-Transformers (SETs) \cite{fuchs2020se} and Steerable Equivariant Graph Neural Networks (SEGNNs) \cite{brandstetter2022geometric}, both of which have previously been evaluated for their ability to model 3D molecular systems. We evaluate these models using default hyperparameters specified by the authors of each method. Regarding QA, besides including all geometric deep learning methods trained to predict per-residue LDDT-C$\alpha$ scores as baselines, we also compare \textsc{EGR}-AllAtom and \textsc{EGR}-C$\alpha$ to GNN\_DOVE \cite{10.3389/fmolb.2021.647915}, a state-of-the-art (SOTA) graph deep learning method for all-atom protein complex structure QA.\par

\textbf{\textsc{EGR} Models.} In addition to our original \textsc{EGR} model (i.e., \textsc{EGR}-AllAtom), we include three separate versions of the \textsc{EGR} model where we individually remove node or edge-specific features from our input graphs during training and test time. These versions are the \textsc{EGR} model without positional corruption (i.e., \textsc{EGR}-AllAtom-NPC), the model without atom-wise protein surface proximities (i.e., \textsc{EGR}-AllAtom-NSP), and the model without edge-wise relative geometric features (i.e., \textsc{EGR}-AllAtom-NRGF). Such versions are listed in Appendix \ref{sec:appendix_a} to serve as ablation studies on the \textsc{EGR} model, to understand the relative importance of individual input features in our datasets.\par

\textbf{Evaluation Metrics.} For the task of refinement, for all relevant metrics we report the change in the metric's value (i.e., the $\Delta$ metric) when comparing a model's refined structure and the original decoy structure. To evaluate the structural quality of a protein complex, we use the DockQ score \cite{basu2016dockq}, interface root mean squared deviation (iRMSD), ligand RMSD (LRMSD), fraction of decoy DockQ scores improved (FI-DockQ), and average percentage of decoy DockQ score improvement (API-DockQ). Here, iRMSD is the mean squared error (MSE) between the atoms of the predicted and ground truth inter-chain interaction regions, and LRMSD is the MSE between the atoms of the predicted and ground truth ligand chains, respectively. Additionally, $F_{nat}$ is the fraction of native contacts found in a decoy structure. Formally, DockQ score can be calculated using Equations \ref{eq:dockq_rmsd} and \ref{eq:dockq_main}, letting $d_{1}=8.5$ and $d_{2}=1.5$:

\begin{equation}
\label{eq:dockq_rmsd}
    RMSD_{\text {scaled }}\left(RMSD, d_{i}\right)=\frac{1}{1+\left(\frac{RMSD}{d_{i}}\right)^{2}}
\end{equation}

\begin{equation}
\label{eq:dockq_main}
    \text{DockQ}=\frac{1}{3} \left(F_{\text {nat }}+ RMSD_{\text {scaled }}\left(LRMSD, d_{1}\right) + RMSD_{\text {scaled }}\left(iRMSD, d_{2}\right)\right).
\end{equation}

To evaluate a model's ranking ability, we use two common metrics, its Top-$N$ hit rate and ranking loss \cite{simpkin2021evaluation, lensink2021prediction}. A model's hit rate is defined as the fraction of protein target complexes where the model ranked at least one acceptable or higher quality decoy within its predicted list of Top-$N$ decoys. A hit rate is represented by three numbers separated by the character /. The three numbers, in order, represent how many decoys with acceptable or higher quality, medium or higher quality, and high quality were among the Top-$N$ ranked decoys. In this work, we employ the Top-10 hit rate measure. Similarly, a per-target ranking loss is defined as the difference between the DockQ score of a target's native structure and the decoy for which the model predicted the highest structural quality. As such, a low ranking loss reflects a model's ability to successfully rank decoys for downstream tasks.

\textbf{Implementation Details.} We train all our models using the AdamW optimizer \cite{loshchilov2017decoupled} and perform early stopping with a patience of 50 epochs based on the average full-complex RMSD of a model's predictions on our PSR dataset's validation split. All hyperparameters and protein graph features employed are described in Appendix \ref{sec:appendix_b}. Source code to perform fast structural refinement and quality assessment using our provided model weights can be found at \href{https://github.com/BioinfoMachineLearning/DeepRefine}{\texttt{https://github.com/BioinfoMachineLearning/DeepRefine}}.

\subsection{Results}
\label{sec:experiments_results}

\begin{table}
    \caption{Performance of different refinement methods on each test dataset.}
    \label{tbl:refinement_performance}
    \centering
    \scriptsize  % Or footnotesize, scriptsize, tiny, etc.
    \begin{tabular}{lccccccc}
        \toprule
        $\Delta$Metric             & DockQ\ $\uparrow$   & iRMSD\ $\downarrow$    & LRMSD\ $\downarrow$     &      FI-DockQ\ $\uparrow$  &     API-DockQ\ $\uparrow$     \\ \midrule
                                   &                   &                       & \underline{PSR-Dockground (4,799)}             &                          &    \\ \midrule
        Modeller                   & +0.0002  & -0.6331   & -1.0027     &       63.03\%    &       0.32\%       \\ \midrule
        \textsc{EGR}-C$\alpha$-Modeller               & +0.0053\ $\pm$\ 0.0011 & -1.2285\ $\pm$\ 0.0330   & -3.5226\ $\pm$\ 0.3125     &       79.30\%\ $\pm$\ 0.93\%  &       0.89\%\ $\pm$\ 0.15\%       \\ \midrule
        SET-AllAtom               & +0.0132\ $\pm$\ 0.0040 & -0.8808\ $\pm$\ 0.1158   & -1.6478\ $\pm$\ 0.1047     &       84.90\%\ $\pm$\ 1.13\%  &       1.69\%\ $\pm$\ 0.35\%       \\ \midrule
        SEGNN-AllAtom               & +\textbf{0.0144}\ $\pm$\ \textbf{0.0024}  & -\textbf{2.4562}\ $\pm$\ \textbf{0.0499}   & -\textbf{6.6603}\ $\pm$\ \textbf{0.6702}     &       \textbf{94.46}\%\ $\pm$\ \textbf{0.60}\%  &       1.89\%\ $\pm$\ 0.29\%       \\ \midrule
        \underline{\textsc{EGR}-AllAtom}               & +0.0097\ $\pm$\ 0.0002 & -0.6274\ $\pm$\ 0.0669   & -2.5561\ $\pm$\ 0.1584     &       83.66\%\ $\pm$\ 0.49\%  &       1.59\%\ $\pm$\ 0.11\%       \\ \midrule
                                   &                   &                       & \underline{PSR-DeepHomo (376)}               &                          &    \\ \midrule
        Modeller                   & -0.2465 & +1.5912   & +5.3457     &       8.24\%  &       0.53\%        \\ \midrule
        \textsc{EGR}-C$\alpha$-Modeller               & -0.2796\ $\pm$\ 0.0055  & +2.2075\ $\pm$\ 0.0839   & +6.1711\ $\pm$\ 0.1842     &       8.16\%\ $\pm$\ 0.76\%  &       1.17\%\ $\pm$\ 0.18\%       \\ \midrule
        SET-AllAtom               & -0.0034\ $\pm$\ 0.0003 & +0.0275\ $\pm$\ 0.0050   & +0.0273\ $\pm$\ 0.0104     &       27.39\%\ $\pm$\ 4.36\%  &       0.20\%\ $\pm$\ 0.08\%       \\ \midrule
        SEGNN-AllAtom               & -0.0468\ $\pm$\ 0.0091 & +0.2950\ $\pm$\ 0.0741   & +0.3593\ $\pm$\ 0.1722     &       16.31\%\ $\pm$\ 3.54\%  &       0.87\%\ $\pm$\ 0.20\%       \\ \midrule
        \underline{\textsc{EGR}-AllAtom}               & -\textbf{0.0006}\ $\pm$\ \textbf{0.0018} & +\textbf{0.0121}\ $\pm$\ \textbf{0.0054}   & +\textbf{0.0013}\ $\pm$\ \textbf{0.0028}     &       \textbf{45.12}\%\ $\pm$\ \textbf{6.99}\%  &       0.41\%\ $\pm$\ 0.03\%       \\ \midrule
                                   &                   &                       & \underline{PSR-EVCoupling (195)}             &                          &    \\ \midrule
        Modeller                   & -0.1738  & +1.1467   & +4.9877     &       7.18\%  &       0.74\%        \\ \midrule
        \textsc{EGR}-C$\alpha$-Modeller               & -0.2150\ $\pm$\ 0.0073  & +1.9651\ $\pm$\ 0.0647   & +5.8477\ $\pm$\ 0.7759     &       9.91\%\ $\pm$\ 1.74\%  &       1.49\%\ $\pm$\ 0.37\%       \\ \midrule
        SET-AllAtom               & -0.0016\ $\pm$\ 0.0002 & +\textbf{0.0149}\ $\pm$\ \textbf{0.0007}   & +0.0108\ $\pm$\ 0.0040     &       27.86\%\ $\pm$\ 5.24\%  &       0.31\%\ $\pm$\ 0.11\%       \\ \midrule
        SEGNN-AllAtom               & -0.0250\ $\pm$\ 0.0069  & +0.1646\ $\pm$\ 0.0633   & +0.2400\ $\pm$\ 0.1044     &       18.29\%\ $\pm$\ 3.41\%  &       0.89\%\ $\pm$\ 0.18\%       \\ \midrule
        \underline{\textsc{EGR}-AllAtom}               & +\textbf{0.0010}\ $\pm$\ \textbf{0.0010} & +0.0026\ $\pm$\ 0.0031   & -\textbf{0.0059}\ $\pm$\ \textbf{0.0017}     &       \textbf{43.93}\%\ $\pm$\ \textbf{5.00}\%  &       0.48\%\ $\pm$\ 0.03\%       \\ \midrule
                                   &                   &                       & \underline{Benchmark 2 (17)}            &                          &    \\ \midrule
        Modeller                   & -0.1855  & +0.7939   & +3.0277     &       5.88\%  &       0.60\%        \\ \midrule
        GalaxyRefineComplex               & -0.0074  & +0.0778   & -\textbf{0.0246}     &       22.22\%  &       2.12\%       \\ \midrule
        GNNRefine               & +0.0025  & +0.0226   & +0.0602     &       47.06\%  &       1.26\%       \\ \midrule
        \textsc{EGR}-C$\alpha$-Modeller               & -0.2644\ $\pm$\ 0.0437 & +2.118\ $\pm$\ 0.7832   & +5.9196\ $\pm$\ 1.8589     &       15.69\%\ $\pm$\ 2.77\%  &       1.28\%\ $\pm$\ 0.84\%       \\ \midrule
        SET-AllAtom               & -0.0078\ $\pm$\ 0.0015 & +0.0729\ $\pm$\ 0.0186   & +0.0469\ $\pm$\ 0.0114     &       29.63\%\ $\pm$\ 2.62\%  &       0.33\%\ $\pm$\ 0.14\%       \\ \midrule
        SEGNN-AllAtom               & -0.0328\ $\pm$\ 0.0062  & +0.0807\ $\pm$\ 0.0790   & +0.0781\ $\pm$\ 0.1371     &       31.37\%\ $\pm$\ 5.54\%  &       1.24\%\ $\pm$\ 0.59\%       \\ \midrule
        \underline{\textsc{EGR}-AllAtom}               & -\textbf{0.0010}\ $\pm$\ \textbf{0.0028} & -\textbf{0.0002}\ $\pm$\ \textbf{0.003}   & -0.0121\ $\pm$\ 0.0021     &       \textbf{43.14}\%\ $\pm$\ \textbf{10.00}\%  &       0.59\%\ $\pm$\ 0.08\%       \\ \midrule
    \end{tabular}
\end{table}

\textbf{Blind Refinement.} In Table \ref{tbl:refinement_performance}, we see that the all-atom \textsc{EGR} model surpasses the refinement performance of all other baseline methods except on the low initial quality structures from Dockground. We observe that methods outperforming \textsc{EGR} on the Dockground targets also appear to underperform \textsc{EGR} on targets from datasets of higher initial structural quality, suggesting these methods are overfitting to low-quality structures in the PSR training dataset. \textsc{EGR}-AllAtom, however, appears to generalize better to all test datasets compared to any other method (accounting for the fact that GNNRefine may have seen overly-similar proteins during training), with a computational complexity (e.g., average of 5 seconds) orders of magnitude lower than that of methods such as GalaxyRefineComplex and GNNRefine (e.g., average of 1,200 and 600 seconds, respectively).\par

\begin{table}
    \caption{Hit rate performance of different QA methods on the M4S test dataset.}
    \label{tbl:m4s_ranking_performance}
    \centering
    \scriptsize  % Or footnotesize, scriptsize, tiny, etc.
    \begin{tabular}{lcccccc}
        \toprule
        ID      & \textsc{EGR}-C$\alpha$-Modeller  & SET-AllAtom      & SEGNN-AllAtom   & \underline{\textsc{EGR}-AllAtom}    & GNN\_DOVE & \textbf{Top-10 Best} \\ \midrule
        7AOH    & 10/10/6  & 9/8/6     & 9/9/9     & 9/9/9     & 9/9/0              & 10/10/10    \\ \midrule
        7D7F    & 0/0/0     & 2/0/0     & 0/0/0     & 0/0/0     & 0/0/0              & 5/0/0       \\ \midrule
        7AMV    & 10/10/8  & 10/10/5   & 10/10/9   & 10/10/5   & 10/10/6                        & 10/10/10    \\ \midrule
        7OEL    & 10/10/0   & 10/10/0   & 10/9/0    & 10/9/0    & 10/10/0                        & 10/10/0     \\ \midrule
        7O28    & 10/10/0   & 10/10/0   & 10/10/0   & 10/10/0   & 10/10/0                        & 10/10/0     \\ \midrule
        7MRW    & 6/5/0     & 0/0/0     & 0/0/0     & 0/0/0     & 0/0/0                          & 10/10/0     \\ \midrule
        7D3Y    & 0/0/0     & 0/0/0     & 0/0/0     & 1/0/0     & 0/0/0                          & 10/0/0      \\ \midrule
        7NKZ    & 10/10/9  & 10/9/9    & 10/10/3   & 10/9/9    & 10/9/9                         & 10/10/10    \\ \midrule
        7LXT    & 10/10/0     & 4/3/0     & 6/5/0     & 8/7/0     & 1/0/0                          & 10/10/0     \\ \midrule
        7KBR    & 10/10/10   & 10/10/10  & 10/10/10  & 10/10/9  & 10/10/9                        & 10/10/10    \\ \midrule
        7O27    & 10/5/0   & 10/7/0    & 10/6/0    & 10/4/0    & 10/4/0                         & 10/10/0     \\ \midrule
        Summary & \textbf{9}/\textbf{9}/\textbf{4}   & \textbf{9}/8/\textbf{4}    & 8/8/\textbf{4} & \textbf{9}/8/\textbf{4}   & 8/7/3                          & 11/9/4      \\ \bottomrule
    \end{tabular}
\end{table}

\begin{table}
    \caption{Ranking loss of different QA methods on the M4S test dataset.}
    \label{tbl:m4s_ranking_loss}
    \centering
    \scriptsize  % Or footnotesize, scriptsize, tiny, etc.
    \begin{tabular}{lccccc}
        \toprule
        ID      & \textsc{EGR}-C$\alpha$-Modeller  & SET-AllAtom    & SEGNN-AllAtom & \underline{\textsc{EGR}-AllAtom}       & GNN\_DOVE \\ \midrule
        7AOH    & 0.0610    & 0.9280   & 0.9280 & 0.0350     & 0.9280    \\ \midrule
        7D7F    & 0.4700    & 0.4700   & 0.4710 & 0.4590     & 0.0030    \\ \midrule
        7AMV    & 0.1730    & 0.3420   & 0.0130 & 0.3420     & 0.3420    \\ \midrule
        7OEL    & 0.2100    & 0.2100   & 0.3790 & 0.2100     & 0.2100    \\ \midrule
        7O28    & 0.2330    & 0.0240   & 0.2740 & 0.2440     & 0.2440    \\ \midrule
        7MRW    & 0.6000    & 0.5550   & 0.6030 & 0.5550     & 0.5980    \\ \midrule
        7D3Y    & 0.3240    & 0.2950   & 0.1740 & 0.2950    & 0.2950    \\ \midrule
        7NKZ    & 0.0220    & 0.1100   & 0.1830 & 0.4590     & 0.4590    \\ \midrule
        7LXT    & 0.0500    & 0.2950   & 0.2950 & 0.3890     & 0.2950    \\ \midrule
        7KBR    & 0.1700    & 0.1520   & 0.0520 & 0.1520     & 0.0680    \\ \midrule
        7O27    & 0.3340    & 0.3340   & 0.3650 & 0.3180     & 0.3340    \\ \midrule
        Summary & \textbf{0.2406}\ $\pm$\ \textbf{0.1801}   & 0.3377\ $\pm$\ 0.2486 & 0.3397\ $\pm$\ 0.2613   & \textbf{0.3144}\ $\pm$\ \textbf{0.1506}    & 0.3432\ $\pm$\ 0.2538 \\ \bottomrule
    \end{tabular}
\end{table}

\textbf{Blind Quality Assessment.} Table \ref{tbl:m4s_ranking_performance} summarizes a few key findings regarding models' QA performance. The first is that both \textsc{EGR}-AllAtom and \textsc{EGR}-C$\alpha$-Modeller outperform GNN\_DOVE, the state-of-the-art QA predictor for protein complexes, on a large collection of decoy structures. Interestingly, we also find that modeling protein complexes at the granularity of C$\alpha$ atoms leads to \textsc{EGR}-C$\alpha$-Modeller achieving new SOTA results. This suggests interesting avenues for future research into molecular modeling for structural QA. Nonetheless, within a single model, \textsc{EGR}-AllAtom can outperform all other methods for QA while simultaneously achieving SOTA performance for refinement, demonstrating the complementarity of the two tasks in the context of all-atom modeling.

\begin{figure}
    \centering
    \begin{subfigure}{0.31\textwidth}
        \centering
        \textbf{Original Decoy}
        \includegraphics[width=\textwidth]{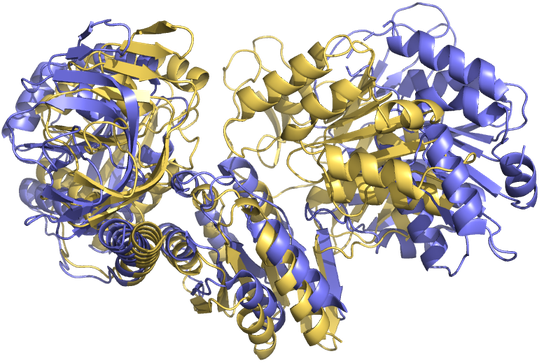}
        \caption*{RMSD = 8.715}
    \end{subfigure}
    \begin{subfigure}{0.31\textwidth}
        \centering
        \textbf{Modeller-Refined Decoy}
        \includegraphics[width=\textwidth]{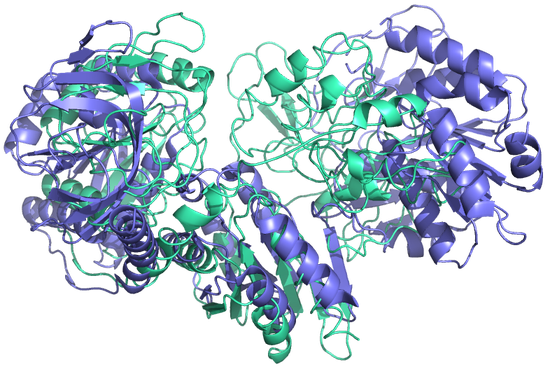}
        \caption*{RMSD = 10.131}
    \end{subfigure}
    \begin{subfigure}{0.31\textwidth}
        \centering
        \textbf{\textsc{EGR}-Refined Decoy}
        \includegraphics[width=\textwidth]{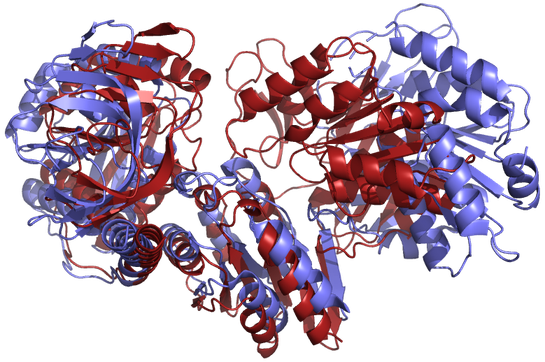}
        \caption*{RMSD = 8.636}
    \end{subfigure}
    \caption{A cherry-picked example of a test protein complex (PDB ID: 6GS2) successfully refined by \textsc{EGR}. Note in all the above subfigures the ground truth structure is highlighted in the color \textsc{slate}.}
    \label{fig:viz_6GS2}
\end{figure}

\textbf{Visualizations.} We display in Figure \ref{fig:viz_6GS2} a successful example of a PSR test protein other refinement methods (e.g., Modeller) cannot refine but that \textsc{EGR} nonetheless can refine atomistically.\par

\textbf{Limitations.} Although \textsc{EGR} has driven significant progress in protein complex structure refinement and assessment, the model is currently only able to improve the structure of inter-chain interfaces marginally on average due partly to its relatively small training dataset size and partly to limited computing resources available for generating sufficiently large datasets. As such, we conjecture that new large datasets comprised of medium and high-quality complex decoys are required for refinement methods to better generalize their predictions to higher-quality structures. We leave an exploration of such ideas for future work.

\section{Conclusion}
\label{sec:conclusion}
We presented \textsc{EGR} which introduces an E(3)-equivariant method for refining and assessing the quality of 3D protein complex structures. Our experiments validate the effectiveness of \textsc{EGR} for structural refinement and assessment using a diverse, open-source collection of protein complexes and establish a baseline for future studies in geometric deep learning for protein structure refinement and analysis. Used with care and with feedback from domain experts, we believe \textsc{EGR} could encourage the use of DL models in responsible drug discovery and development.

\section{Acknowledgments}
The project is partially supported by two NSF grants (DBI 1759934 and IIS 1763246), one NIH grant (GM093123), three DOE grants (DE-SC0020400, DE-AR0001213, and DE-SC0021303), and the computing allocation on the Summit compute cluster provided by Oak Ridge Leadership Computing Facility (Contract No. DE-AC05-00OR22725).

{
\small
\printbibliography
}

\appendix

\section{Additional Results}
\label{sec:appendix_a}

\begin{table}[H]
    \caption{Ablation studies on the \textsc{EGR} model using each test dataset.}
    \label{tbl:refinement_ablation_studies}
    \centering
    \scriptsize  % Or footnotesize, scriptsize, tiny, etc.
    \begin{tabular}{lccccccc}
        \toprule
        $\Delta$Metric             & DockQ\ $\uparrow$   & iRMSD\ $\downarrow$    & LRMSD\ $\downarrow$     &      FI-DockQ\ $\uparrow$  &     API-DockQ\ $\uparrow$     \\ \midrule
                                   &                   &                       & \underline{PSR-Dockground (4,799)}             &                          &    \\ \midrule
        \textsc{EGR}-AllAtom-NPC               & +0.0083\ $\pm$\ 0.0006 & -0.6549\ $\pm$\ 0.1589   & -\textbf{2.5129}\ $\pm$\ \textbf{0.1584}     &       82.95\%\ $\pm$\ 1.98\%  &       1.46\%\ $\pm$\ 0.09\%       \\ \midrule
        \textsc{EGR}-AllAtom-NSP               & +0.0048\ $\pm$\ 0.0039 & -0.4881\ $\pm$\ 0.1622   & -1.5799\ $\pm$\ 0.5891     &       74.15\%\ $\pm$\ 1.77\%  &       1.46\%\ $\pm$\ 0.09\%       \\ \midrule
        \textsc{EGR}-AllAtom-NRGF               & +0.0035\ $\pm$\ 0.0018 & -0.4817\ $\pm$\ 0.0470   & -0.8711\ $\pm$\ 0.2320     &       73.61\%\ $\pm$\ 2.39\%  &       0.71\%\ $\pm$\ 0.19\%       \\ \midrule
        \underline{\textsc{EGR}-AllAtom}               & +\textbf{0.0097}\ $\pm$\ \textbf{0.0002} & -\textbf{0.6274}\ $\pm$\ \textbf{0.0669}   & -2.5561\ $\pm$\ 0.2893     &       \textbf{83.66}\%\ $\pm$\ \textbf{0.49}\%  &       1.59\%\ $\pm$\ 0.11\%       \\ \midrule
                                   &                   &                       & \underline{PSR-DeepHomo (376)}               &                          &    \\ \midrule
        \textsc{EGR}-AllAtom-NPC               & -0.0011\ $\pm$\ 0.0019 & +\textbf{0.0044}\ $\pm$\ \textbf{0.0021}   & -\textbf{0.0003}\ $\pm$\ \textbf{0.0010}     &       36.08\%\ $\pm$\ 9.57\%  &       0.30\%\ $\pm$\ 0.08\%       \\ \midrule
        \textsc{EGR}-AllAtom-NSP               & \textbf{0.0000}\ $\pm$\ \textbf{0.0015} & +0.0088\ $\pm$\ 0.0041   & +0.0008\ $\pm$\ 0.0015     &       46.10\%\ $\pm$\ 11.47\%  &       0.39\%\ $\pm$\ 0.10\%       \\ \midrule
        \textsc{EGR}-AllAtom-NRGF               & -0.1296\ $\pm$\ 0.0276 & +0.7784\ $\pm$\ 0.1660   & +0.5925\ $\pm$\ 0.1625     &       10.28\%\ $\pm$\ 3.83\%  &       1.11\%\ $\pm$\ 0.16\%       \\ \midrule
        \underline{\textsc{EGR}-AllAtom}               & -0.0006\ $\pm$\ 0.0018 & +0.0121\ $\pm$\ 0.0054   & +0.0013\ $\pm$\ 0.0028     &       \textbf{45.12}\%\ $\pm$\ \textbf{6.99}\%  &       0.41\%\ $\pm$\ 0.03\%       \\ \midrule
                                   &                   &                       & \underline{PSR-EVCoupling (195)}             &                          &    \\ \midrule
        \textsc{EGR}-AllAtom-NPC               & -0.0003\ $\pm$\ 0.0014 & +\textbf{0.0007}\ $\pm$\ \textbf{0.0009}   & -0.0041\ $\pm$\ 0.0012     &       33.16\%\ $\pm$\ 10.60\%  &       0.31\%\ $\pm$\ 0.08\%       \\ \midrule
        \textsc{EGR}-AllAtom-NSP               & +0.0007\ $\pm$\ 0.0013 & +0.0021\ $\pm$\ 0.0016   & -0.0057\ $\pm$\ 0.0020     &       39.66\%\ $\pm$\ 9.02\%  &       0.39\%\ $\pm$\ 0.13\%       \\ \midrule
        \textsc{EGR}-AllAtom-NRGF               & -0.0743\ $\pm$\ 0.0183 & +0.4843\ $\pm$\ 0.1084   & +0.3218\ $\pm$\ 0.1003     &       11.11\%\ $\pm$\ 4.48\%  &       1.07\%\ $\pm$\ 0.18\%       \\ \midrule
        \underline{\textsc{EGR}-AllAtom}               & +\textbf{0.0010}\ $\pm$\ \textbf{0.0010} & +0.0026\ $\pm$\ 0.0031   & -\textbf{0.0059}\ $\pm$\ \textbf{0.0017}     &       \textbf{43.93}\%\ $\pm$\ \textbf{5.00}\%  &       0.48\%\ $\pm$\ 0.03\%       \\ \midrule
                                   &                   &                       & \underline{Benchmark 2 (17)}            &                          &    \\ \midrule
        \textsc{EGR}-AllAtom-NPC               & -0.0006\ $\pm$\ 0.0014 & -0.0038\ $\pm$\ 0.0008   & -0.0127\ $\pm$\ 0.0067     &       41.18\%\ $\pm$\ 12.71\%  &       0.26\%\ $\pm$\ 0.20\%       \\ \midrule
        \textsc{EGR}-AllAtom-NSP               & -0.0007\ $\pm$\ 0.0030 & -\textbf{0.0049}\ $\pm$\ \textbf{0.0016}   & -\textbf{0.0185}\ $\pm$\ \textbf{0.0020}     &       41.18\%\ $\pm$\ 12.71\%  &       0.51\%\ $\pm$\ 0.18\%       \\ \midrule
        \textsc{EGR}-AllAtom-NRGF               & -0.0070\ $\pm$\ 0.0054 & +0.0010\ $\pm$\ 0.0027   & -0.0126\ $\pm$\ 0.0004     &       27.45\%\ $\pm$\ 7.34\%  &       0.22\%\ $\pm$\ 0.14\%       \\ \midrule
        \underline{\textsc{EGR}-AllAtom}               & -\textbf{0.0010}\ $\pm$\ \textbf{0.0028} & -0.0002\ $\pm$\ 0.003   & -0.0121\ $\pm$\ 0.0021     &       \textbf{43.14}\%\ $\pm$\ \textbf{10.00}\%  &       0.59\%\ $\pm$\ 0.08\%       \\ \midrule
    \end{tabular}
\end{table}

\begin{table}[H]
    \caption{Performance of different refinement methods, including ablation studies, on all test datasets.}
    \label{tbl:psr_dockground_refinement_performance}
    \centering
    \scriptsize  % Or footnotesize, scriptsize, tiny, etc.
    \begin{tabular}{lccccccc}
        \toprule
        $\Delta$Metric                          & Fraction of Native Contacts\ $\uparrow$    &    Fraction of Non-Native Contacts\ $\downarrow$     \\ \midrule
                                                &   & \underline{PSR-Dockground (4,799)}    \\ \midrule
        Modeller                                & +0.0003    & +\textbf{0.0009}  \\ \midrule
        \textsc{EGR}-C$\alpha$-Modeller               & +0.0112\ $\pm$\ 0.0034    & +0.0029\ $\pm$\ 0.0007  \\ \midrule
        SET-AllAtom               & +0.0392\ $\pm$\ 0.0126    & +0.0109\ $\pm$\ 0.0008  \\ \midrule
        SEGNN-AllAtom       & +0.0322\ $\pm$\ 0.0072    & +0.0094\ $\pm$\ 0.0006 \\ \midrule
        \textsc{EGR}-AllAtom-NPC       & +0.0284\ $\pm$\ 0.0024    & +0.0116\ $\pm$\ 0.0001 \\ \midrule
        \textsc{EGR}-AllAtom-NSP       & +0.0153\ $\pm$\ 0.0110    & +0.0085\ $\pm$\ 0.0026 \\ \midrule
        \textsc{EGR}-AllAtom-NRGF       & +0.0128\ $\pm$\ 0.0110    & +0.0098\ $\pm$\ 0.0018 \\ \midrule
        \underline{\textsc{EGR}-AllAtom}               & +\textbf{0.0324}\ $\pm$\ \textbf{0.0027}    & +0.0122\ $\pm$\ 0.0011  \\ \midrule
                                                &   & \underline{PSR-DeepHomo (376)}      \\ \midrule
        Modeller                    & -0.1844   & +0.2445  \\ \midrule
        \textsc{EGR}-C$\alpha$-Modeller               & -0.1630\ $\pm$\ 0.0106    & +0.3441\ $\pm$\ 0.0072  \\ \midrule
        SET-AllAtom               & +0.0014\ $\pm$\ 0.0013    & +0.0144\ $\pm$\ 0.0043  \\ \midrule
        SEGNN-AllAtom                & -0.0152\ $\pm$\ 0.0064    & +0.1032\ $\pm$\ 0.0349  \\ \midrule
        \textsc{EGR}-AllAtom-NPC       & -0.0009\ $\pm$ 0.0063    & +\textbf{0.0029}\ $\pm$ \textbf{0.0069} \\ \midrule
        \textsc{EGR}-AllAtom-NSP       & +0.0049\ $\pm$\ 0.0065    & +0.0143\ $\pm$\ 0.0126 \\ \midrule
        \textsc{EGR}-AllAtom-NRGF       & -0.0942\ $\pm$\ 0.0672    & +0.2930\ $\pm$\ 0.0665 \\ \midrule
        \underline{\textsc{EGR}-AllAtom}               & +\textbf{0.0046}\ $\pm$\ \textbf{0.0029}    & +0.0157\ $\pm$\ 0.0016  \\ \midrule
                                                &   & \underline{PSR-EVCoupling (195)}    \\ \midrule
        Modeller                   & -0.1334    & +0.1971  \\ \midrule
        \textsc{EGR}-C$\alpha$-Modeller               & -0.1343\ $\pm$\ 0.0131    & +0.3100\ $\pm$\ 0.0082  \\ \midrule
        SET-AllAtom               & +0.0020\ $\pm$\ 0.0014    & +0.0110\ $\pm$\ 0.0064  \\ \midrule
        SEGNN-AllAtom               & -0.0005\ $\pm$\ 0.0030    &    +0.0873\ $\pm$\ 0.0315  \\ \midrule
        \textsc{EGR}-AllAtom-NPC       & 0.0003\ $\pm$ 0.0045    & +\textbf{0.0009}\ $\pm$ \textbf{0.0049} \\ \midrule
        \textsc{EGR}-AllAtom-NSP       & +0.0047\ $\pm$\ 0.0049    & +0.0105\ $\pm$\ 0.0086 \\ \midrule
        \textsc{EGR}-AllAtom-NRGF       & -0.0482\ $\pm$\ 0.0475    & +0.2064\ $\pm$\ 0.0555 \\ \midrule
        \underline{\textsc{EGR}-AllAtom}               & +\textbf{0.0065}\ $\pm$\ \textbf{0.0015}    & +0.0113\ $\pm$\ 0.0022  \\ \midrule
                                                &   & \underline{Benchmark 2 (17)}       \\ \midrule
        Modeller                   & -0.1645    & +0.2232       \\ \midrule
        GalaxyRefineComplex         & -0.0001   &   +0.0601     \\ \midrule
        GNNRefine         & +0.0087   &   +0.0707     \\ \midrule
        \textsc{EGR}-C$\alpha$-Modeller               & -0.1794\ $\pm$\ 0.0405    &    +0.3915\ $\pm$\ 0.0432  \\ \midrule
        SET-AllAtom               & +\textbf{0.0072}\ $\pm$\ \textbf{0.0033}    & +0.0212\ $\pm$\ 0.0034  \\ \midrule
        SEGNN-AllAtom               & -0.0275\ $\pm$\ 0.0080    & +0.0781\ $\pm$\ 0.0058  \\ \midrule
        \textsc{EGR}-AllAtom-NPC       & -0.0004\ $\pm$ 0.0067    & 0.0000\ $\pm$ 0.0057 \\ \midrule
        \textsc{EGR}-AllAtom-NSP       & +0.0005\ $\pm$\ 0.0088    & +0.0053\ $\pm$\ 0.0086 \\ \midrule
        \textsc{EGR}-AllAtom-NRGF       & -0.0157\ $\pm$\ 0.0142    & -\textbf{0.0075}\ $\pm$\ \textbf{0.0099} \\ \midrule
        \underline{\textsc{EGR}-AllAtom}               & +0.0014\ $\pm$\ 0.006    & +0.0110\ $\pm$\ 0.0037  \\ \midrule
    \end{tabular}
\end{table}

In Table \ref{tbl:refinement_ablation_studies}, we investigate the effect of removing individual components or input features from the \textsc{EGR} model. We then compare these results to the baseline \textsc{EGR} model to assess these components' relative importance on generalization and model accuracy.
Similarly, in addition to iRMSD, LRMSD, FI-DockQ, and API-DockQ, we can also inspect the fraction of native and non-native inter-chain contacts present in a structural decoy for a protein target. Such measures can yield insights into how the interface between two or more chains in a complex was formed and whether or not the interface is \textit{near-native}. We show the results of these measures in Table \ref{tbl:psr_dockground_refinement_performance}.

\section{Implementation Details}
\label{sec:appendix_b}

\textbf{Expanded discussion of attention mechanism.} As mentioned in Section \ref{sec:egr_model}, within each \textsc{EGR} layer, we compute $a_{i}$ from $\mathbf{H}$ using a variant of the Linear Attention Transformer architecture \cite{vaswani2017attention, DBLP:journals/corr/abs-1812-01243, wang2021}. In particular, we compute not only global attention scores for each atom pair, but we also calculate local, atom-wise attention scores using windows of size $128 \times 128$. To perform global attention, the most computationally-expensive of these operations, in linear time, a key insight from \cite{DBLP:journals/corr/abs-1812-01243} is that\par

\begin{equation}
    \mathbf{D}(\mathbf{Q}, \mathbf{K}, \mathbf{V}) = \frac{\mathbf{Q} \mathbf{K}^{\top}}{n}\mathbf{V} = \mathbf{E}(\mathbf{Q}, \mathbf{K}, \mathbf{V}) = \frac{\mathbf{Q}}{\sqrt{n}}\left( \frac{\mathbf{K}^{\top}}{\sqrt{n}}\mathbf{V} \right)
\end{equation}

where the commutativity of scalar multiplication with matrix multiplication and the associativity of matrix multiplication give us

\begin{equation}
\begin{aligned}
    \mathbf{E}(\mathbf{Q}, \mathbf{K}, \mathbf{V}) & = \frac{\mathbf{Q}}{\sqrt{n}}\left( \frac{\mathbf{K}^{\top}}{\sqrt{n}}\mathbf{V} \right)\\
               & = \frac{1}{n}\mathbf{Q}\left( \mathbf{K}^{\top}\mathbf{V} \right)\\
               & = \frac{1}{n}\left( \mathbf{Q}\mathbf{K}^{\top} \right)\mathbf{V}\\
               & = \frac{\mathbf{Q} \mathbf{K}^{\top}}{n}\mathbf{V},
\end{aligned}
\end{equation}

with $\mathbf{D}$ representing standard attention that is quadratic w.r.t. the input sequence length; $\mathbf{E}$ denoting linear attention that is linear w.r.t. the input sequence length; $\mathbf{Q}$, $\mathbf{K}$, and $\mathbf{V}$, respectively, being query, key, and value linear projections of $\mathbf{H}$; and $n$ representing a normalization constant. As such, our global attention calculations remain computationally feasible for large input protein complexes. Our local attention calculations, then, follow closely after those of \cite{beltagy2020longformer}.\par

\textbf{Featurization.} In each all-atom protein complex graph, we include as a node feature an atom's type using a 38-dimensional one-hot encoding vector for each atom. Then, to add a new node feature representing the normalized proximity of an atom to the surface of the protein chain to which it belongs, using MSMS \cite{sanner1996reduced} we take the complement of each atom's chain-wise buriedness to derive atomic surface proximities for surface contact modeling \cite{ganea2021independent}. Notably, for the \textsc{EGR}-C$\alpha$-Modeller model, its input graphs are comprised of C$\alpha$ atoms as nodes rather than all available atoms. In such C$\alpha$ atom graphs, we also include node features describing each residue's dihedral angles \cite{jin2021iterative}. Lastly, we note that in our experiments with the SEGNN, for this model we included an additional type-1 (i.e., vector-valued) node feature describing the coordinates-wise displacement between a given node's position and the mean atomic coordinates of the protein complex. All node features we selected for all-atom and C$\alpha$-atom graphs are shown in Tables \ref{tab:all_atom_graph_features} and \ref{tab:ca_atom_graph_features}, respectively.\par

For our graphs' edge features, we start by adding a binary feature indicating whether or not a pair of connected atoms belong to the same chain. We then add an edge-wise sinusoidal positional encoding of the difference between the source and destination node's indices in the input graph \cite{morehead2021geometric}. As another edge feature, we include a binary value indicating whether or not a pair of atoms are connected by a covalent bond. Our final type-0 edge feature describes the relative geometric features such as distance, direction, and orientation between the local coordinate systems representing the residues corresponding to an atom pair, following \cite{ingraham2019generative, jumper2021highly}. Lastly, in our SEGNN experiments, we also included an additional type-1 edge feature describing the relative coordinates-wise displacement between a connected pair of atoms. All edge features we selected for all-atom and C$\alpha$-atom graphs are shown in Tables \ref{tab:all_atom_graph_features} and \ref{tab:ca_atom_graph_features}, respectively.\par

\begin{table}[H]
\caption{Summary of \textsc{EGR}'s node and edge features for all-atom graphs. Here, $N$ and $E$ denote the number of nodes and edges in $\mathcal{G}$, respectively.}
\label{tab:all_atom_graph_features}
\centering
\begin{tabular}{clll}
\toprule
                                             & \multicolumn{1}{c}{Feature} & \multicolumn{1}{c}{Type} & Shape \\ \midrule
\multirow{1}{*}{Node Features}               & One-hot encoding of atom type         & Categorical                        & $N \times 38$          \\
                                             & Chain-local surface proximity         & Numeric                            & $N \times 1 $          \\ \midrule
\multirow{1}{*}{Edge Features}               & Permutation-invariant chain encoding  & Categorical                        & $E \times 1 $          \\
                                             & Sinusoidal edge positional encoding   & Numeric                            & $E \times 1 $          \\
                                             & Relative geometric features           & Numeric                            & $E \times 12$          \\
                                             & Covalent bond encoding                & Categorical                        & $E \times 1 $          \\ \midrule
\multicolumn{1}{l}{\multirow{1}{*}{Total}}   & Node features                         &                                    & $N \times 39$          \\
\multicolumn{1}{l}{}                         & Edge features                         &                                    & $E \times 15$          \\ \bottomrule
\end{tabular}
\end{table}

\begin{table}[H]
\caption{Summary of \textsc{EGR}'s node and edge features for C$\alpha$-atom graphs. Here, $N$ and $E$ denote the number of nodes and edges in $\mathcal{G}$, respectively.}
\label{tab:ca_atom_graph_features}
\centering
\begin{tabular}{clll}
\toprule
                                             & \multicolumn{1}{c}{Feature} & \multicolumn{1}{c}{Type} & Shape \\ \midrule
\multirow{1}{*}{Node Features}               & One-hot encoding of residue type      & Categorical                        & $N \times 21$          \\
                                             & Chain-local surface proximity         & Numeric                            & $N \times 1 $          \\
                                             & Dihedral angle descriptors            & Numeric                            & $N \times 6 $          \\ \midrule
\multirow{1}{*}{Edge Features}               & Permutation-invariant chain encoding  & Categorical                        & $E \times 1 $          \\
                                             & Sinusoidal edge positional encoding   & Numeric                            & $E \times 1 $          \\
                                             & Relative geometric features           & Numeric                            & $E \times 12$          \\ \midrule
\multicolumn{1}{l}{\multirow{1}{*}{Total}}   & Node features                         &                                    & $N \times 28$          \\
\multicolumn{1}{l}{}                         & Edge features                         &                                    & $E \times 14$          \\ \bottomrule
\end{tabular}
\end{table}

\textbf{Hardware Used.} The Oak Ridge Leadership Facility (OLCF) at the Oak Ridge National Laboratory (ORNL) is an open science computing facility that supports HPC research. The OLCF houses the Summit compute cluster. Summit, launched in 2018, delivers 8 times the computational performance of Titan’s 18,688 nodes, using only 4,608 nodes. Like Titan, Summit has a hybrid architecture, and each node contains multiple IBM POWER9 CPUs and NVIDIA Volta GPUs all connected with NVIDIA’s high-speed NVLink. Each node has over half a terabyte of coherent memory (high bandwidth memory + DDR4) addressable by all CPUs and GPUs plus 800GB of non-volatile RAM that can be used as a burst buffer or as extended memory. To provide a high rate of I/O throughput, the nodes are connected in a non-blocking fat-tree using a dual-rail Mellanox EDR InfiniBand interconnect. We used the Summit compute cluster to train all our models.\par

\textbf{Software Used.} We used Python 3.8 \cite{10.5555/1593511}, PyTorch 1.10.0 \cite{NEURIPS2019_9015}, PyTorch Lightning 1.5.10 \cite{falcon2019pytorch}, PyTorch Geometric 2.0.4 \cite{Fey_Fast_Graph_Representation_2019}, and DGL 0.8 \cite{wang2019dgl} to run our deep learning experiments. PyTorch Lightning was used to facilitate model checkpointing, metrics reporting, and distributed data parallelism across 144 Tesla V100 GPUs. A more in-depth description of the software environment used to run inference with our models can be found at \href{https://github.com/BioinfoMachineLearning/DeepRefine}{\texttt{https://github.com/BioinfoMachineLearning/DeepRefine}}.\par

\textbf{Further hyperparameters.} We use a learning rate of $10^{-4}$ for all \textsc{EGR} models. The learning rate is kept constant throughout each model's training. Models with the lowest RMSD on our validation data split are then tested on all our sequence-filtered test splits.\par

\begin{table}[H]
    \caption{Hyperparameter search space for all \textsc{EGR} models through which we searched to obtain strong performance on the PSR validation split. The final parameters for the standard \textsc{EGR} model are in \textbf{bold}.}
    \label{tbl:hyperparameter_search_space}
    \centering
    % \scriptsize  % Or footnotesize, scriptsize, tiny, etc.
    \begin{tabular}{lc}
        \toprule
        Hyperparameter   &   Search Space \\ \midrule
        PSR Loss Weight   &   1.0 (Based on Loss on Validation Split) \\
        QA Loss Weight   &   \textbf{0.05}, 0.005 \\
        Number of Layers   &   4, \textbf{7} \\
        Hidden Dimension    &  \textbf{64}, 128 \\
        Non-Linearities    &  \textbf{LeakyReLU}, PReLU, SiLU \\
        Learning Rate   & \textbf{0.0001}, 0.001 \\
        Weight Decay Rate   & $10^{-8}$, $\mathbf{10^{-4}}$ \\
        Normalization   & \textbf{LayerNorm}, GroupNorm, Identity \\
        \bottomrule
    \end{tabular}
\end{table}

\end{document}